\documentclass{article}
\usepackage{dingbat}
\usepackage{spconf,amsmath,graphicx}
\usepackage{amsfonts}
\usepackage{xcolor}
\usepackage{hyperref}
\usepackage{xspace}\usepackage[compact]{titlesec}         % you need this package
\title{ABCDE: An Agent-Based Cognitive Development Environment}
\name{Jieyi Ye$^{1}$ \qquad Jiafei Duan$^{2}$ \qquad Samson Yu$^{3}$ \qquad Bihan Wen$^{1}$ \qquad Cheston Tan$^{3}$}

\address{\{ye003yi, bihan.wen\}@e.ntu.edu.sg,
\{duan\_jiafei, cheston-tan\}@i2r.a-star.edu.sg, samson yu@ihpc.a-star.edu.sg\\
$^{1}$Nanyang Technological University, Singapore\\
$^{2}$Institute for Infocomm Research, A*STAR\\
$^{3}$Centre for Frontier AI Research \\
}

\begin{document}
% \titlespacing{\section}{0pt}{0pt}{0pt}
\setlength\abovedisplayskip{0pt}
\setlength\belowdisplayskip{0pt}
%\ninept
%
\maketitle
\begin{abstract}
Children's cognitive abilities are sometimes cited as AI benchmarks. How can the most common 1,000 concepts (89\% of everyday use) be learnt in a naturalistic children's setting? Cognitive development in children is about quality, and new concepts can be conveyed via simple examples. Our approach of knowledge scaffolding uses simple objects and actions to convey concepts, like how children are taught. We introduce ABCDE, an interactive 3D environment modeled after a typical playroom for children. It comes with 300+ unique 3D object assets (mostly toys), and a large action space for child and parent agents to interact with objects and each other. ABCDE is the first environment aimed at mimicking a naturalistic setting for cognitive development in children; no other environment focuses on high-level concept learning through learner-teacher interactions. The simulator can be found at \url{https://pypi.org/project/ABCDESim/1.0.0/}
 
\end{abstract}
\section{Introduction}
\label{sec:intro}

The cognitive abilities of children are sometimes held up as informal benchmarks for AI. For example, Bengio in 2019 said that “I don't think we're anywhere close today to the level of intelligence of a two-year-old child”~\cite{bengio}. Typical 5-year-olds have probably experienced between 0.5M to 50M\footnote{5.3M = 5 years x 365 days x 12 waking hours x 240 ``learning events'' (each 15 secs) per hour. Then allow for factor of 10 more or fewer.} “training examples” (including many repetitions), which are not too far off from the size of current AI datasets, e.g. 22M images~\cite{deng2009imagenet} and 8M videos~\cite{youtube}.

However, the characteristics are very different. 5-year-olds are estimated to recognize 5K to 10K words~\cite{Howmany} -- far fewer than the 20K+ object classes in ImageNet and 100K+ synsets in WordNet. Given a hypothetical “budget” of 5M training examples, children would experience something like 5K classes x 20 unique instances x 50 variations, rather than a typical AI dataset of say 20K classes x 250 unique instances. Of course, children eventually attain adult-level vocabularies, typically 25K to 50K English words~\cite{Howmany}. 

Cognitive development is about quality before quantity. The current AI paradigm of ``training'' is akin to animal training. However, conceptual knowledge can be conveyed via simplified, abstracted examples~\cite{zitnick2013bringing}. Children are taught, not trained, and educational interactions are designed to convey ideas with clarity and illustrativeness.

In this work, we introduce ABCDE, an interactive 3D environment modeled after a typical playroom for children, along with child and parent avatars. It comes with 300+ unique 3D object assets, most being toys with various physical attributes. There is a large action space for child and parent agents to interact with objects and each other. ABCDE is the first interactive environment that mimics a naturalistic setting for cognitive development in children; no other environment \cite{puig2018virtualhome,duan2022survey,kolve2017ai2,savva2019habitat,gan2020threedworld} focuses on high-level concept learning through learner-teacher interactions.
% Furthermore, we offer a comprehensive set of actions, observations and metadata of the scene via our Python module to enable a wide range of machine learning tasks
\begin{figure}[ht]
    \centering
    \includegraphics[width=\linewidth]{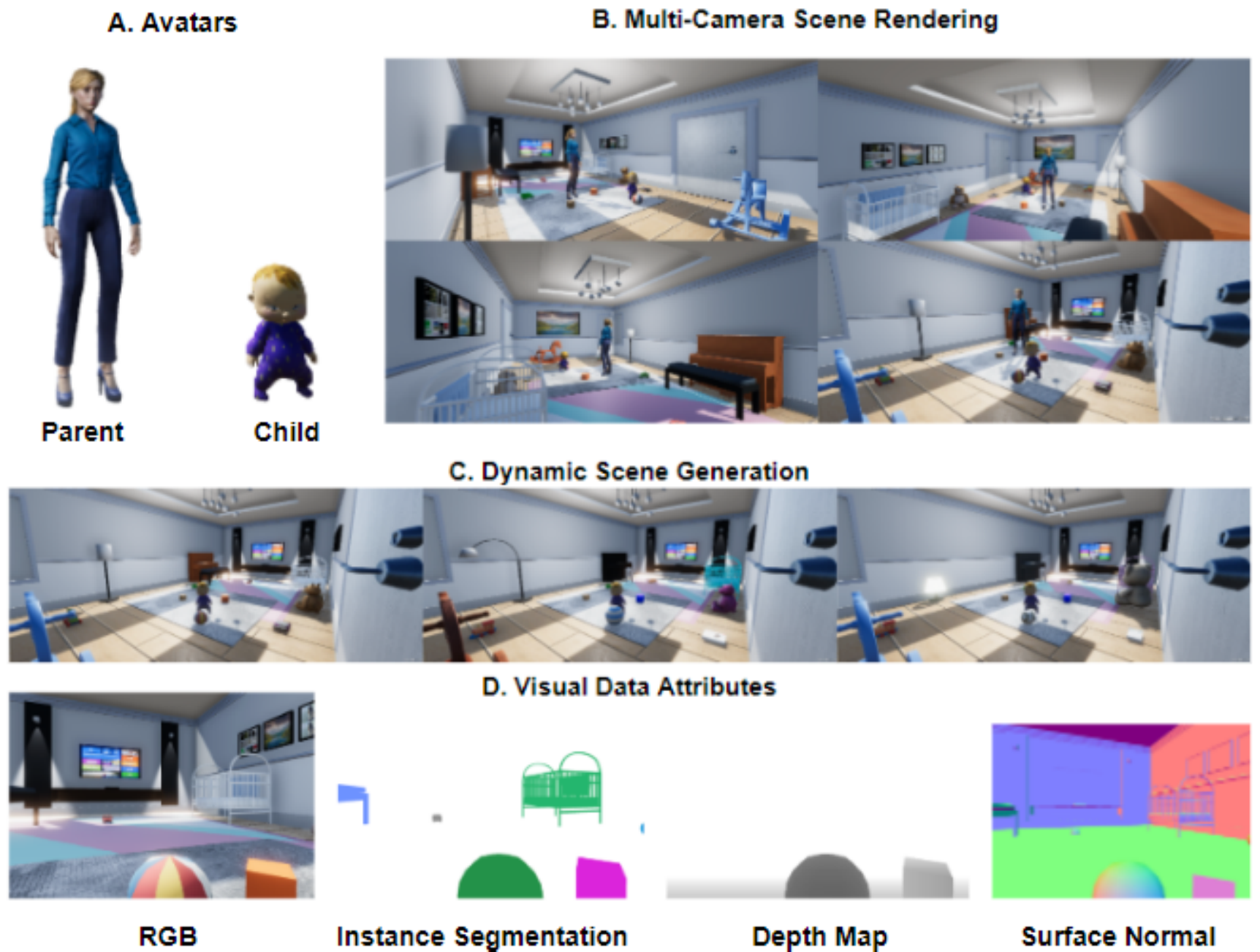}
    
    \caption{Overview of the various components of ABCDE.}
\label{fig:1}
\end{figure}

\begin{figure}[ht]
    \centering
    \includegraphics[width=\linewidth]{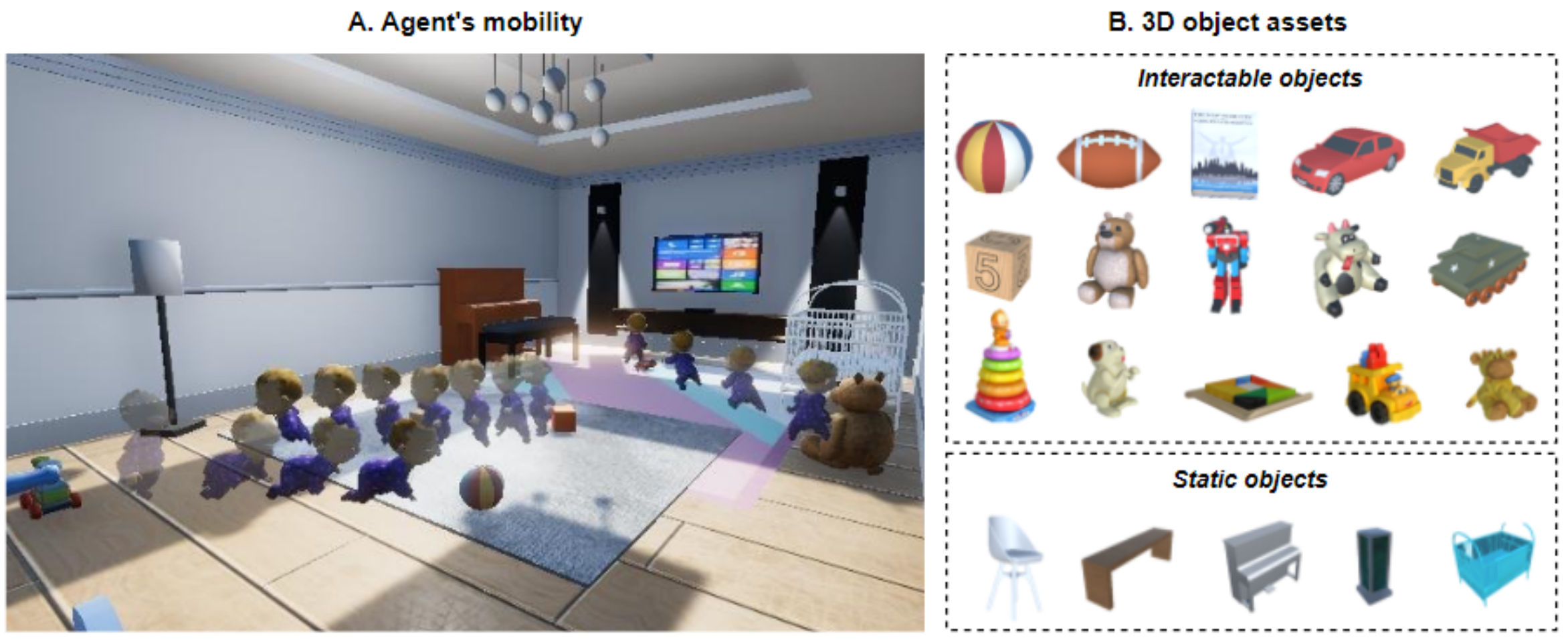}
    
    \caption{A) Time-lapse visualization depicting some child actions. B) The two categories of 3D object assets.}
\label{fig:2}
\end{figure}

%------------------------------------------------------------------------
\section{Objective and Approach}

The objective of developing ABCDE is to start addressing this challenge: Can common concepts be learnt in a grounded, naturalistic children's setting? The most common 1,000 concepts cover 89\% of everyday usage~\cite{Howmany}. They fall into a variety of word classes; not just nouns, verb and adjectives, but also other important classes such as determiners, prepositions, etc. In contrast, AI tasks have mostly been for nouns (objects), verbs (actions) and adjectives (attributes), and these have mostly been tackled in isolation.

Our proposed approach is to use simple objects and actions to convey concepts, similar to how young children are taught in a playroom. This approach uses knowledge scaffolding, e.g. using tangible nouns and verbs to convey abstract prepositions and determiners, etc. The focus is on common concepts, not necessarily activities of adult daily living (e.g. cooking and general household tasks) \cite{shridhar2020alfred,damen2018scaling,duan2020actionet}. A parent avatar is used for realism (objects can’t move by themselves), embodiment, referencing (e.g. pointing to an object being mentioned verbally) and for imitation learning (i.e. for child agents to observe body motions). ABCDE can be used for both training and evaluation of learning agents.

%------------------------------------------------------------------------
\section{ABCDE Simulator}

The Agent-Based Cognitive Development Environment (ABCDE) is designed with 3 key aspects. \textbf{1)} A parent (teacher) agent, which can interact with the child (learner) agent in several ways, including passively ``speaking'' (generating natural language text) when the child is playing, and also actively demonstrating to the child. \textbf{2)} The natural language text ``spoken'' by the parent can be generated at different levels of grammatical complexity. \textbf{3)} The parent can be algorithmically programmed to carry out high-level actions to demonstrate concepts, meaning that labelled video examples can be generated for free. For instance, one way to demonstrate the ``put-on'' concept is to pick up object A, then lower it over object B until they are in contact (e.g. ``put the cup on the table''). This approach allows for tangible demonstration for grounding of natural language concepts that can sometimes be abstract, e.g. ``\textbf{only} the red ball is in the box, not the green balls''.

\subsection{Scene Structure}
ABCDE contains a single scene that allows for an unlimited amount of room configuration through the random sampling of interactable objects at potential locations on the floor grid. Figure \ref{fig:1}C shows the dynamic generation of scene. The interactable objects will be spawned with a separation buffer of +/-\emph{0.5} units between objects, so as to prevent overlaps. More object classes can be easily imported and spawned at available locations on the grid. Dynamic generation could also be initialized via the Python module. All the objects' positions and metadata are logged.

\subsection{Actions}
The simulator allows both the child and parent agents to interact with and move around the environment via specific actions. For navigation, the available actions available are [\emph{NavigateTo, Run, Crawl, WalkForward, WalkBackwards, TurnLeft, TurnRight, Wander}]. For interactions, the actions are [\emph{Grab, LookAt, PutBack, LookAround, Touch, Rotate}]. The interactions with the object assets within the environment are also influenced by the internal proprieties of the 3D objects. The 3D objects in the environment are split into two categories: [\emph{Static, Interactable}]. All interactions use the Nvidia PhysX Engine to enable realistic physics.

\subsection{Observations}
ABCDE supports a wide range of visual data attributes, from RGB, depth and instance segmentation to surface normal data (Figure \ref{fig:1}D). We provide four stationary cameras to provide full visual coverage of the environment and render all four cameras simultaneously (Figure \ref{fig:1}B). All observations are output at customizable frame rates. We also provide a GUI for easier control of the scene and agents.

\subsection{3D Object Assets}
We use 332 3D object assets from public marketplaces and the Google Scanned Object dataset \cite{Downs2022GoogleSO}, falling into two categories: 89 static objects and 243 interactable objects.

\subsection{Tasks}

%Our goal in building this simulator is to provide a realistic and configurable virtual environment. We provide a functional simulator and a programmable python module. We are eager to curate a set of tasks to evaluate learning.

Following the nature of different types of knowledge and the corresponding organization of human memory~\cite{tan2021comprehensive, tan2017vision}, the tasks cover both declarative and procedural concepts. Declarative concepts include ``static'' concepts such as nouns, adjectives, etc.
%ACBDE proposes to take a relational or comparative approach to tasks. For instance, the concepts of “big” and “small” are highly contextual (a big mouse is smaller than a small elephant), but using the algorithmic approach in ABCDE, comparative demonstrations of a “big [X]” next to a “small [X]” can be generated effortlessly.
Procedural concepts are typically actions or action sequences.
%For example, the “keep in box” concept might be composed of “take”, then “put into box”, then “close box”.
These concepts could potentially be tested in various ways, including telling the child agent to demonstrate the concept (e.g. ``\textbf{keep} the red ball in the box''), Q\&A (e.g. ``is the toy car \textbf{on} the table or \textbf{under} it?''), fill-in-the-blank (``the toy car is \rule{0.5cm}{0.15mm} the table''), etc.

%------------------------------------------------------------------------
% \section{Final copy}

% You must include your signed IEEE copyright release form when you submit your finished paper.
% We MUST have this form before your paper can be published in the proceedings.

% Please direct any questions to the production editor in charge of these proceedings at the IEEE Computer Society Press:
% \url{https://www.computer.org/about/contact}.

%%%%%%%%% REFERENCES

\bibliographystyle{IEEEbib}
\bibliography{strings,refs}

\end{document}